\documentclass[letterpaper, 10 pt, journal, twoside]{IEEEtran}
\IEEEoverridecommandlockouts
\usepackage{amsmath,amssymb, graphicx, epstopdf,cite,enumerate,booktabs, setspace, float,stfloats,bm, multirow, lscape, array, bbding}
\usepackage[table]{xcolor}
\definecolor{purple2}{HTML}{EEE6FD}

\usepackage[normalem]{ulem}

\allowdisplaybreaks[4]
\usepackage{caption}
\usepackage{subcaption}
\usepackage{graphbox}
\usepackage{soul, color, xcolor}
\usepackage[bottom=1.52cm,top=2.01cm,left=1.69cm,right=1.69cm]{geometry}

\usepackage[linesnumbered,ruled]{algorithm2e}

\vspace{-0.1in}
\title{\LARGE \bf Self-Evolutionary Replanning for Failure-Aware Motion Planning}

\author{Guoliang Li$^{1}$, Ruihua Han$^{3}$, Chengyang Li$^{3}$, He Li$^{1}$, Shuai Wang$^{2,\dagger}$,\\Wenchao Ding$^{4}$, Hong Zhang$^{5}$,~\emph{Life Fellow, IEEE}, Chengzhong Xu$^{1,\dagger}$,~\emph{Fellow, IEEE}
\vspace{-0.3in}
\thanks{$^\dagger$ denotes the corresponding authors: s.wang@siat.ac.cn, czxu@um.edu.mo}
\thanks{$^{1}$ Department of Computer and Information Science, University of Macau}
\thanks{$^{2}$ SIAT, Chinese Academy of Sciences}
\thanks{$^{3}$ Department of Computer Science, University of Hong Kong}
\thanks{$^{4}$ Academy for Engineering \& Technology, Fudan University}
\thanks{$^{5}$ Department of EEE, Southern University of Science and Technology}
}

\begin{document}
\maketitle

\begin{abstract}
Motion planners tuned under nominal conditions can fail when execution conditions change. Existing differentiable adaptation methods provide efficient online updates but remain constrained by static loss functions and gradient-only parameter updates. We propose Self-Evolutionary RePlanning (SERP), a failure-aware framework that jointly adapts motion-planner parameters and loss function weights from closed-loop execution feedback. SERP provides two adaptation mechanisms. First, in-context learning with auto-differentiation (ILAD) enables rapid online adaptation by coupling local numerical refinement with history-conditioned semantic intervention. Auto-differentiation performs numerical updates, while a vision-language model (VLM) interprets recent adaptation history to jointly revise planner parameters and loss weights when local adaptation becomes ineffective. Second, the memory-augmented rollout system (MARS) extends ILAD into a deliberative and evolutionary recovery process with memory-guided candidate generation and parallel low-fidelity rollout evaluation. The resulting recovery trace is then consolidated into recovery memory to guide future adaptation. Extensive simulation and real-world experiments demonstrate improved failure recovery performance over various baselines across diverse environments.


\end{abstract}

\section{Introduction} \label{section1} 

Robots operating in complex environments inevitably encounter failures that cannot be anticipated during planner design, where environmental uncertainties may render preset parameters ineffective. To
address this issue, replanning (RP) \cite{gao2020teach,han2025ral,wang2024flexible} is a promising
solution, where the robot is allowed to fail, but is capable of adjusting plan until success. Furthermore, an execution failure may be caused by not only external environment uncertainties, but also ego model uncertainties.

Recent differentiable RP methods address the model uncertainties by adapting its internal model via closed-loop execution loss backpropagation \cite{agrawal2019differentiable,tao2024difftune,cheng2024difftune,han2025neupan}.
However, static loss functions and gradient-only parameter updates may become ineffective when navigation conditions change. Importantly, unsuccessful adaptation also provides useful evidence: the evolution of planner parameters together with the resulting
closed-loop outcomes can reveal why the current optimization direction
fails. Language model based methods have shown that semantic feedback can be used to revise optimization objectives or subsequent decisions \cite{ma2024eureka,shinn2023reflexion}. This capability provides a complementary means of reshaping planner adaptation when numerical refinement becomes ineffective.

\begin{figure}[t!]
\centering
\includegraphics[width=0.45\textwidth]{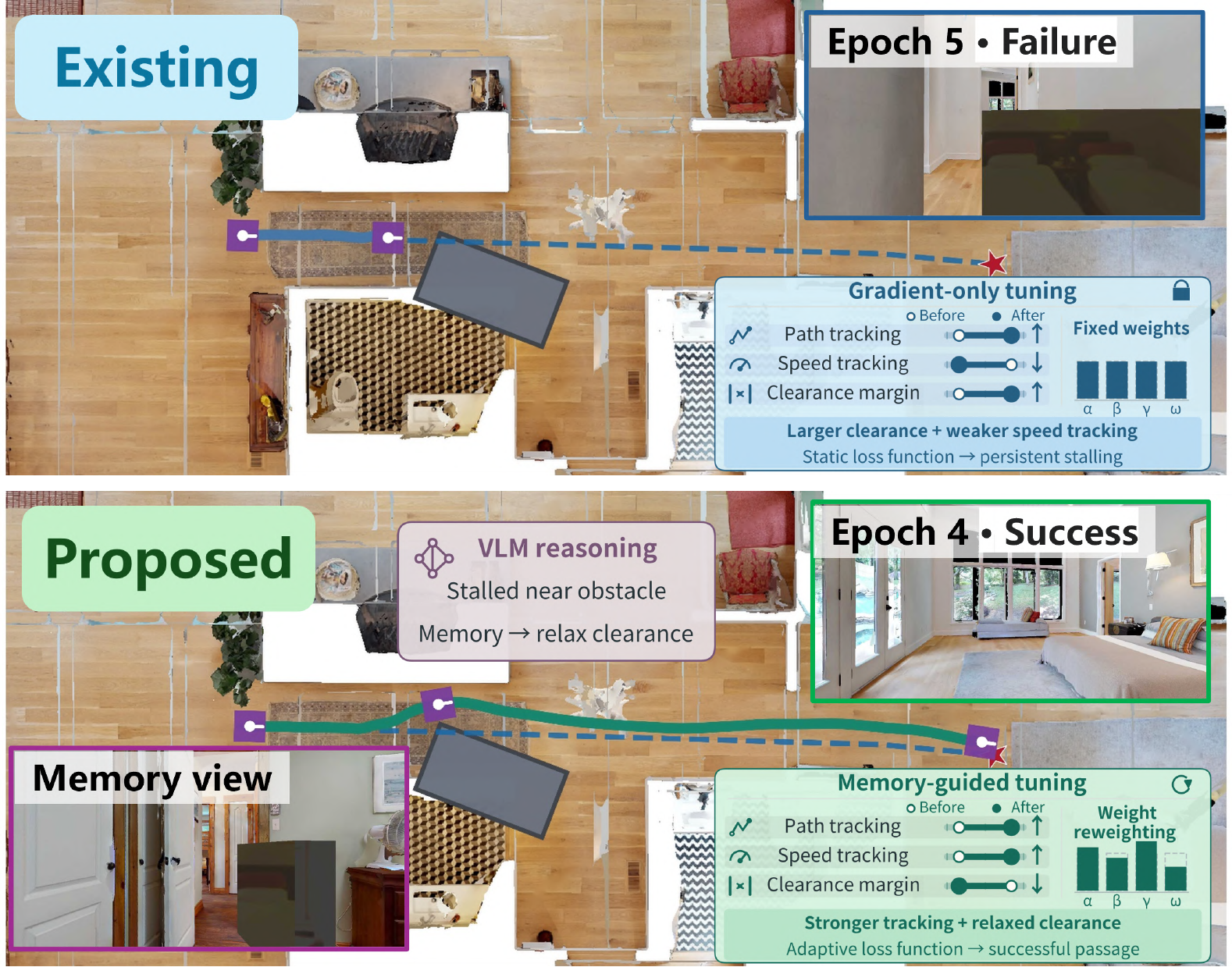}
\caption{Comparison of planner adaptation methods.}
\label{fig1}
\end{figure}

Beyond a single planner adaptation path, failure recovery may benefit from exploring multiple candidate strategies and assessing their closed-loop consequences before execution. Directly testing
these candidates in the target environment, however, may introduce
additional execution cost and safety risk. Moreover, previous embodied-agent studies
have demonstrated the value of retaining and reusing interaction
experience for subsequent reasoning and decision making
\cite{anwar2025remembr,zhao2024expel,wang2023voyager}. For planner
adaptation, accumulated recovery experience can similarly guide the
construction of alternative configurations, while
lightweight closed-loop rollouts provide behavioral evidence for
identifying a promising recovery direction.

Motivated by these observations, we propose Self-Evolutionary
RePlanning (SERP), a failure-aware framework that jointly
adapts planner parameters and loss weights from closed-loop execution feedback. SERP provides two adaptation mechanisms. First, in-context learning with auto-differentiation (ILAD)
integrates numerical refinement with semantic intervention conditioned on recent adaptation history. Second, the memory-augmented rollout system
(MARS) extends the single semantic intervention to a deliberative and evolutionary process with memory-guided candidate generation, parallel rollout evaluation, and recovery memory accumulation. Fig.~\ref{fig1} illustrates how SERP successfully recovers from failure by leveraging both reasoning and memory, whereas gradient-only tuning with static loss function remains ineffective.

\begin{figure*}[!t]
\centering
\includegraphics[width=0.95\textwidth]{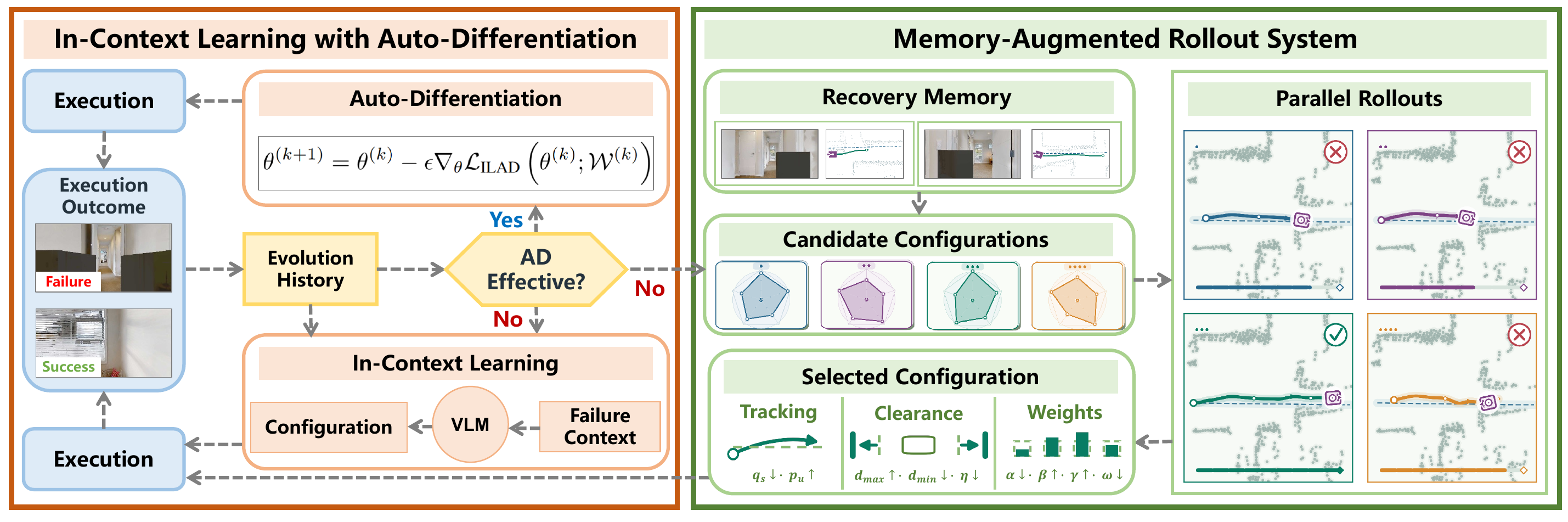}
\caption{Overview of SERP, which consists of ILAD and MARS.}
\label{fig2}
\vspace{-0.2in}
\end{figure*}

Our contributions are summarized as follows:
\begin{itemize}
    \item We introduce SERP, a self-evolutionary replanning
    framework that jointly adapts planner parameters and loss weights from closed-loop execution feedback.

    \item We develop ILAD, which integrates auto-differentiation based numerical refinement with history-conditioned semantic intervention to enable rapid online adaptation.

    \item We further develop MARS, which extends ILAD with
    memory-guided multiple candidate generation, low-fidelity rollout
evaluation and memory accumulation.

    \item We evaluate SERP in IR-SIM, Habitat, and real-world experiments, demonstrating the superior failure-recovery performance of SERP.
\end{itemize}

\section{Related Work}\label{section2}

\subsection{Replanning and Differentiable Planner Adaptation}

Robot replanning (RP) revises navigation behavior when environmental conditions or execution outcomes deviate from prior assumptions. 
Optimization-based methods regenerate or refine feasible motions under newly observed constraints
\cite{gao2020teach,zhou2021raptor,wang2024flexible,
zhang2020optimization,han2023rda}, while learning-based approaches improve motion generation from data
\cite{qureshi2020motion,han2025ral}. Existing RP methods typically respond to newly acquired environmental information by revising the planned trajectory or motion. Despite these advancements, existing RP methods freeze the ego action model, thereby missing opportunities to identify potentially better plans during the replanning process.

To enhance RP, enabling robot to adapt its internal model is a promising direction, so as to avoid
repeated failures in future executions. Differentiable optimization layers and differentiable MPC enable gradients to propagate through optimization-based decision processes
\cite{agrawal2019differentiable,amos2018differentiable}, which provides a direct mechanism for internal model adaptation. This principle has been applied to closed-loop controller and MPC tuning \cite{cheng2024difftune,tao2024difftune}, as well as differentiable navigation planners \cite{ishida2022towards,han2025neupan}. Although these methods support efficient gradient-based optimization, their updates are typically governed by static loss functions and gradient-only parameter updates. ILAD instead uses recent adaptation history to reshape subsequent gradient updates by jointly revising planner parameters and loss weights when local refinement becomes ineffective.

\subsection{Reasoning and Memory Guided Adaptation}

Large language models (LLMs) and vision-language models (VLMs) are increasingly used to adapt optimization objectives and decision processes in response to execution feedback. They have been applied to generate and refine reward functions and training configurations \cite{xie2024text2reward,ma2024eureka,ma2024dreureka}. Beyond objective design, LLMs can also reason about execution failures and revise task plans. DEPS uses an LLM to describe plan execution, explain failures, and revise previously generated plans
\cite{wang2023deps}, whereas LERa uses a VLM to interpret visual feedback, diagnose execution errors, and trigger replanning
\cite{pchelintsev2025lera}. ILAD is related to these approaches but operates at a different level: instead of revising a high-level task plan, it integrates VLM-based reasoning into differentiable planner adaptation to reshape the adaptation process when local gradient-based refinement becomes ineffective.

Memory-based approaches provide a complementary means of reusing
knowledge acquired across interactions. HELPER retrieves previously stored planning and correction examples to support subsequent execution and failure recovery
\cite{sarch2023helper}, whereas ExpeL distills previous trajectories into reusable experience for future decisions
\cite{zhao2024expel}. 
MSI-Agent further organizes past experience into multi-scale insights for retrieval
\cite{fu2024msi}, and JARVIS-1 maintains multimodal interaction memory to support planning and continual improvement
\cite{wang2025jarvis1}. 
In robot navigation, ReMEmbR stores long-horizon visual and spatio-temporal observations for later reasoning
\cite{anwar2025remembr}. 
These approaches mainly reuse plans, observations, semantic insights, or task-level experience. Unlike these approaches, MARS accumulates planner recovery memory by retaining successful and unsuccessful adaptation traces together with their closed-loop outcomes. Retrieved memory further guides the candidate configuration generation, which can be used to identify the effective adaptation direction via parallel low-fidelity rollouts.

\section{System Overview}\label{section3}

The architecture of SERP is illustrated in
Fig.~\ref{fig2}. SERP studies how closed-loop execution feedback can be used to adapt the system configuration $\mathcal{Q}=\left[\mathcal{\theta},\mathcal{W}\right]$, where $\mathcal{\theta}$ and $\mathcal{W}$ are the planner parameters and loss weights, respectively. Consider a navigation planner $\pi$ parameterized by $\mathcal{\theta}$, which generates an executable state-action sequence $\mathcal{A}$ according to a navigation plan $\mathcal{S}$ and physical constraints $\mathcal{C}$. Formally,
\begin{align}
\mathcal{A}
=
\pi(\mathcal{\theta},\mathcal{S},\mathcal{C}).
\end{align}
Executing $\mathcal{A}$ in the environment produces an outcome
\begin{align}\label{execution_outcome}
\mathcal{O}
=
\texttt{Execute}(\mathcal{A},\mathcal{C}).
\end{align}

Although the navigation plan $\mathcal{S}$ may remain feasible, the conditions encountered during execution can differ from those assumed during planner design. 
As a result, the current configuration may become unsuitable for the encountered condition, leading to an undesirable outcome, e.g., collision. Upon failure, the robot rolls back to a prior feasible state along the executed trajectory for subsequent execution.

At adaptation epoch $k$, let $\mathcal{X}_k$ denote the current
failure context, comprising visual observation and execution outcome. In addition,
$\mathcal{B}_k$ records the recent adaptation history for $\mathcal{Q}$ and corresponding closed-loop consequences:
\begin{align}\label{adaptation_history}
\mathcal{B}_k
=
\left\{
\left(
\mathcal{Q}^{(i)},
\mathcal{L}^{(i)},
\mathcal{O}^{(i)}
\right)
\right\},
\end{align}
where $\mathcal{L}^{(i)}$ is the corresponding adaptation loss and $\mathcal{O}^{(i)}$ represents the resulting closed-loop outcome. 
The history $\mathcal{B}_k$ captures how previous parameter and loss weight updates affect the resulting navigation behavior.

We formulate replanning as an adaptation operator $\Phi_r$ that updates the configuration according to $\mathcal{X}_k$ and $\mathcal{B}_k$:
\begin{align}\label{problem}
\mathcal{Q}^{(k+1)}=
\Phi_r
\left(
\mathcal{X}_k,
\mathcal{B}_k,
\mathcal{Q}^{(k)}
\right).
\end{align}
The objective is to obtain an updated configuration that recovers from the current failure while maintaining feasible closed-loop execution.

\section{Self-Evolutionary Replanning}\label{section4}

We next present the two adaptation mechanisms of SERP. 
ILAD performs efficient local planner adaptation through auto-differentiation and invokes in-context learning only when recent gradient-based adaptation becomes ineffective. 
MARS further extends the semantic intervention of ILAD with memory-guided candidate generation and parallel low-fidelity rollout evaluation.

\subsection{In-Context Learning with Auto-Differentiation}\label{section5}

Given a state-action sequence $\mathcal{A}
=\{(\mathbf{a}_t,\mathbf{s}_t),\ldots,
(\mathbf{a}_{t+H},\\\mathbf{s}_{t+H})\}$, the planner generates actions by solving
\begin{align}\label{PL}
\mathcal{A}(\mathcal{\theta})
=
\underset{\mathcal{A}\in\mathcal{A}_{\text{feas}}}
{\operatorname{arg\,min}}
\;
F(\mathcal{A}\mid\mathcal{\theta}),
\end{align}
where $\mathcal{A}_{\text{feas}}$ denotes the feasible action space.
Over the receding horizon
$\mathcal{H}=\{t,\ldots,t+H\}$,
the planning objective is given by
\cite{han2025neupan,zhang2020optimization}
\begin{align}\label{planning_objective}
F(\mathcal{A}\mid\mathcal{\theta})
=
\sum_{h=t}^{t+H}
\Big(
&q_s
\|\mathbf{s}_h-\mathbf{s}^{\diamond}_{h}\|_2^2
+
p_u
\|\mathbf{e}_1^T\mathbf{a}_{h}-v^\diamond\|_2^2
\nonumber\\
&-
\eta
\|\mathbf{d}_h\|_1
\Big).
\end{align}
Here, $\mathbf{s}^{\diamond}_{h}$ and $v^\diamond$ denote the reference state and reference velocity, respectively, while $\mathbf{d}_h$ describes the adaptive safety distance at planning
step $h$, constrained by
$d_{\min} \preceq \mathbf{d}_h \preceq d_{\max}$.
The parameters $d_{\min}$ and $d_{\max}$ define the lower and
upper bounds of the safety distance, respectively. The vector $\mathbf{e}_1=[1,0]^{T}$ selects the linear-velocity component of
$\mathbf{a}_h$. The planner parameters
$\mathcal{\theta}
=\{q_s,p_u,\eta,d_{\min},d_{\max}\}$
jointly determine the tracking and clearance behavior of the planner.

To evaluate the effectiveness of a candidate configuration, ILAD defines an execution-oriented adaptation loss
\begin{align}\label{loss}
\mathcal{L}_{\text{ILAD}}
(\mathcal{\theta};\mathcal{W})
=&
\sum_{h=t}^{t+H}
\Big(
\alpha
\|\mathbf{s}_h(\mathcal{\theta})
-\mathbf{s}^{\diamond}_{h}\|_2^2
+
\beta
\|\mathbf{e}_1^T\mathbf{a}_{h}(\mathcal{\theta})
-v^\diamond\|_2^2
\nonumber\\
&-
\gamma
\|\mathbf{d}_h(\mathcal{\theta})\|_1
\Big)
+
\omega
\|\mathbf{s}_{t+H}(\mathcal{\theta})
-\mathbf{s}_{\mathrm{goal}}\|_2 ,
\end{align}
where
$\mathcal{W}=(\alpha,\beta,\gamma,\omega)$,
$\alpha,\beta,\gamma,\omega\geq0$, and
$\alpha+\beta+\gamma+\omega=1$.
The first three terms evaluate the tracking--avoidance trade-off associated with \eqref{planning_objective}, while the terminal term explicitly measures progress toward the navigation goal.
Since
$\{\mathbf{s}_h,\mathbf{a}_h,\mathbf{d}_h\}$
are obtained through \eqref{PL}, they are implicit functions of $\mathcal{\theta}$, allowing gradients of the execution-oriented loss to propagate through the planner optimization.

Auto-differentiation constitutes the numerical refinement mode of ILAD. At adaptation epoch $k$, when the recent updates remain effective, the planner parameters are updated as
\begin{align}\label{AD}
\mathcal{\theta}^{(k+1)}
=
\mathcal{\theta}^{(k)}
-
\epsilon
\nabla_{\mathcal{\theta}}
\mathcal{L}_{\text{ILAD}}
\left(
\mathcal{\theta}^{(k)};
\mathcal{W}^{(k)}
\right),
\end{align}
while the loss weights remain unchanged,
$\mathcal{W}^{(k+1)}=\mathcal{W}^{(k)}$.
Here, $\epsilon$ denotes the learning rate.
This update exploits the differentiable structure of the planner to efficiently refine continuous parameters using local gradient information.

However, successive AD epochs may remain confined to an ineffective region of the parameter space, particularly when the current loss function trade-off inadequately characterizes the encountered failure.
ILAD therefore monitors the recent adaptation history $\mathcal{B}_k$ defined in Sec.~\ref{section3}.
Rather than treating unsuccessful gradient updates as discarded adaptation steps, ILAD uses this adaptation trajectory as evidence for determining whether continued local optimization remains promising. At each adaptation epoch $k$, ILAD determines whether to continue local AD adaptation or invoke in-context learning (IL). IL is activated when failures persist over the recent adaptation history and the relative loss reduction remains below a threshold $\delta_L$:
\begin{align}\label{trigger}
z_k
=
\mathbb{I}
\left[
\texttt{Fail}(\mathcal{B}_k)
\wedge
\frac{
\mathcal{L}_{\mathrm{first}}
-
\mathcal{L}_{\mathrm{last}}
}{
\max(
|\mathcal{L}_{\mathrm{first}}|,
\varepsilon)
}
<
\delta_L
\right],
\end{align}
where
$\mathcal{L}_{\mathrm{first}}$ and
$\mathcal{L}_{\mathrm{last}}$
denote the first and last adaptation losses in $\mathcal{B}_k$, respectively, and
$\varepsilon$ is a small constant for numerical stability.
When $z_k=0$, the recent gradient adaptation is considered effective and epoch $k$ performs an AD update.
When $z_k=1$, persistent failure together with insufficient loss improvement indicates that continued local adaptation is unlikely to provide effective recovery, and epoch $k$ instead performs an IL update. The complete ILAD update is therefore expressed as
\begin{align}\label{IL}
\left[
\mathcal{\theta}^{(k+1)},~
\mathcal{W}^{(k+1)}
\right]
=
\begin{cases}
\mathsf{VLM}
\left(
\mathcal{X}_k,
\mathcal{B}_k,
\mathcal{\theta}^{(k)},
\mathcal{W}^{(k)}
\right),
& z_k=1,
\\[2mm]
\left[\text{Eq.}~(\ref{AD}),~\mathcal{W}^{(k)}\right], 
& z_k=0. \nonumber
\end{cases}
\end{align}
When $z_k=1$, the VLM jointly interprets $\mathcal{X}_k$ and $\mathcal{B}_k$ to reason why the preceding gradient adaptation has been ineffective.
It then revises both the planner parameters and the loss weights, which reshapes subsequent gradient updates and provides a new initialization for subsequent numerical refinement. After that, $\mathcal B_k$ is also reset. The above procedure is iterated until a feasible solution is
attained, or the maximum replanning epoch threshold ($\mathcal{Y}$) is exceeded, i.e., $k=\mathcal{Y}+1$.

\subsection{Memory-Augmented Rollout System}\label{section6}

ILAD adapts the current system configuration by combining
numerical refinement with semantic intervention along a single adaptation path, where recent adaptation history supports online intervention but is not retained beyond the current recovery process as reusable experience. MARS therefore extends this single semantic revision into a deliberative and evolutionary recovery process. It retrieves recovery memory to generate multiple candidate configurations, evaluates their potential closed-loop consequences through parallel low-fidelity rollouts, and consolidates the resulting recovery trace into memory for future adaptation.

MARS maintains recovery memory $\mathcal{M}$ as
\begin{align}
\mathcal{M}
=
\left\{
\mathcal{M}^{+},
\mathcal{M}^{-}
\right\},
\end{align}
where $\mathcal{M}^{+}$ and $\mathcal{M}^{-}$ store successful and
unsuccessful adaptation traces, respectively. Given the latest failure,
MARS retrieves relevant cases from $\mathcal{M}$. Each failure is indexed
by a first-person RGB image and two fixed-size geometric schematics
centered at the failure position: one preserves the global map axes,
whereas the other is aligned with the local reference path. These
schematics summarize the reference path, executed motion, and surrounding
obstacles. MARS uses \texttt{Qwen3-VL-Embedding-2B} to jointly encode
the RGB observation and geometric schematics into a shared feature
vector for initial similarity-based retrieval. The recalled cases are
then evaluated by \texttt{Qwen3-VL-Reranker-2B}, which compares their
multimodal representations and observable local geometry, including the
path turning angle, minimum clearance, clearance asymmetry, and failure
progress. The highest-ranked eligible cases form the retrieved memory
$\mathcal{R}_k$.

Then the VLM generates $J$ candidate configurations:
\begin{align}\label{mars_candidates}
\hat{\mathcal{Q}}_k
&=
\mathsf{VLM}_{\mathrm{gen}}
\left(
\mathcal{X}_k,
\mathcal{B}_k,
\mathcal{Q}^{(k)},
\mathcal{R}_k
\right)
\nonumber\\
&=
\left\{
\hat{\mathcal{Q}}_{k,1},
\ldots,
\hat{\mathcal{Q}}_{k,J}
\right\},
\end{align}
Direct evaluation of $\hat{\mathcal{Q}}k$ in the target environment would introduce additional execution cost and safety risk. MARS therefore constructs a low-fidelity replay environment using the recorded local perception and reference-path information of the current failure, and evaluates the candidates in parallel. For each candidate, the initial state is $\hat{\theta}_{k,j}^{(0)}=\theta_{k,j}$. If the current rollout does not recover from the failure, AD updates the planner parameters according to Eq.~(\ref{AD}) for subsequent evaluation. This process continues for at most $N$ rollouts and terminates early once successful recovery is observed.

For candidate $j$, the rollout trace is represented as
\begin{align}\label{mars_rollout_trace}
\mathcal{T}_{k,j}
=
\left\{
\left(
\hat{\mathcal{\theta}}_{k,j}^{(n)},
\mathcal{O}_{k,j}^{(n)}
\right)
\right\},
\qquad
0\leq n< N,
\end{align}
where $\mathcal{O}_{k,j}^{(n)}$ denotes the corresponding closed-loop outcome. Since candidate evaluations are independent, the $J$ rollout
trajectories are executed in parallel:
\begin{align}\label{mars_rollout}
\left\{
\mathcal{T}_{k,j}
\right\}_{j=1}^{J}
=
\operatorname{ParallelSimRollout}^{N}
\left(
\hat{\mathcal{Q}}_k,
\mathcal{X}_k
\right).
\end{align}

The VLM then selects a promising configuration based on the rollout outcome:
\begin{align}\label{mars_selection}
(j^\star,n^\star)
=
\mathsf{VLM}_{\mathrm{sel}}
\left(
\left\{
\mathcal{T}_{k,j}
\right\}_{j=1}^{J}
\right).
\end{align}
Here, $(j^\star,n^\star)$ identifies the selected candidate and its
corresponding configuration along the rollout trajectory, which is then applied to the target environment:
\begin{align}\label{mars_update}
\left[
\mathcal{\theta}^{(k+1)},
\mathcal{W}^{(k+1)}
\right]
=
\left[
\hat{\mathcal{\theta}}_{k,j^\star}^{(n^\star)},
\hat{\mathcal{W}}_{k,j^\star}
\right].
\end{align}

After execution, the real recovery trace
$\mathcal{T}^{\mathrm{real}}_k$ is consolidated into memory:
\begin{align}
\mathcal{M}_{k+1}
=
\texttt{Consolidate}
\left(
\mathcal{M}_{k},
\mathcal{T}^{\mathrm{real}}_k
\right).
\end{align}
When no eligible recovery memory is available, candidate generation relies on $\mathcal{X}_k$ and $\mathcal{B}_k$; simulated rollouts are used only as decision evidence and are not consolidated into memory.

\section{Experiments and Results}\label{section6}

\begin{figure}[!t]
\centering
\includegraphics[width=0.45\textwidth]{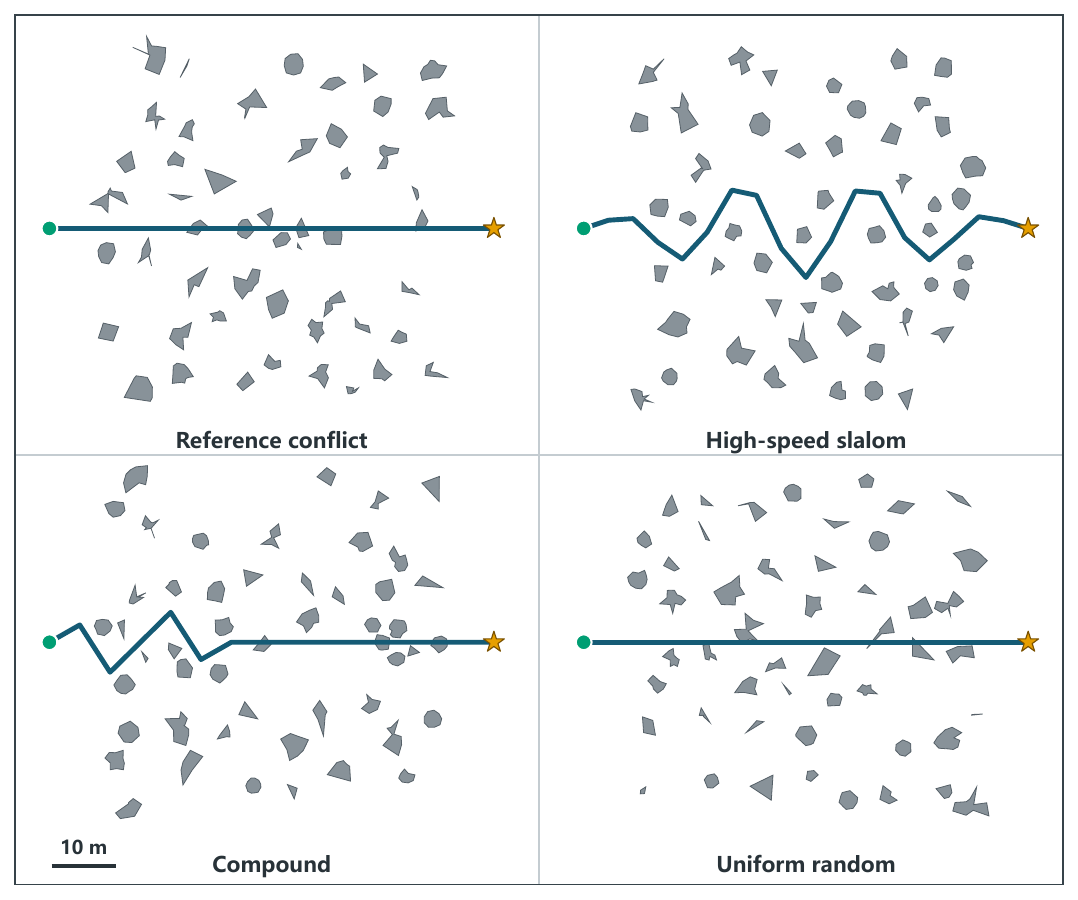}
\caption{IR-SIM scenario profiles.}
\label{fig3}
\end{figure}

We evaluate SERP across three complementary settings. First, IR-SIM \cite{han2026irsim} provides 2D simulation for evaluating the failure-recovery capability of ILAD. Second, Habitat simulator \cite{ramakrishnan2021hm3d} is used to evaluate MARS in more complex indoor environments. Finally, real-world experiments are conducted using a car-like robot in a cluttered laboratory environment to examine the feasibility of SERP under physical execution constraints. All simulation experiments are performed on an Ubuntu workstation equipped with an NVIDIA A6000 GPU.

We compare SERP with four navigation and planner adaptation baselines: SFM \cite{helbing1995social}, ORCA \cite{van2011reciprocal}, NeuPAN \cite{han2025neupan}, and DiffTune \cite{cheng2024difftune}. 
Performance is evaluated using six metrics: 
(a) \textbf{AvgSpeed} (m/s), the mean speed during navigation; 
(b) \textbf{AvgAcc} (m/s$^2$), the mean acceleration magnitude during navigation; (c) \textbf{AvgEpoch}, the average number of adaptation epochs; 
(d) \textbf{TotTime} (s), the average episode duration; 
(e) \textbf{ColRate} (\%), the percentage of episodes terminating with a collision; and 
(f) \textbf{SucRate} (\%), the percentage of episodes in which the robot reaches the goal without collision.

\begin{figure*}[t!]
\centering
\includegraphics[width=0.95\textwidth]{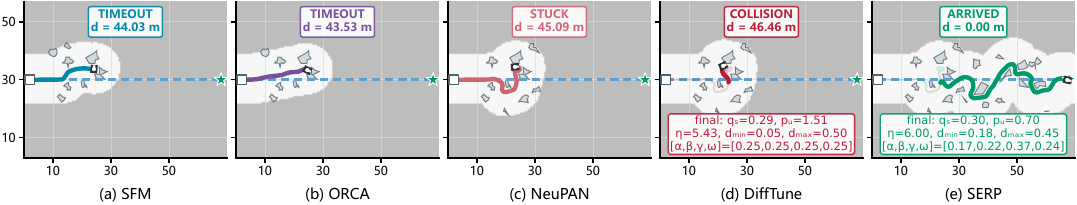} 
\caption{Qualitative comparison in a 50-obstacle uniform-random IR-SIM scene.}
\vspace{-0.03in}
\label{fig4}
\vspace{-0.05in}
\end{figure*}

\subsection{IR-SIM Simulation}

\begin{table}[t] \small
\setlength{\tabcolsep}{1.3mm}
\renewcommand{\arraystretch}{1.32}
\centering
\caption{Quantitative Comparison in the IR-SIM simulator. \colorbox{purple2}{Purple} denotes the \textbf{best} performance.}
\vspace{-0.02in}
\begin{tabular}{cccccc}
\toprule
\multirow{3}{*}{\textbf{Metric}}&\multicolumn{5}{c}{\textbf{Method}}\\
\cmidrule{2-6}
&{SFM}&ORCA&NeuPAN&DiffTune&\multirow{2}{*}{SERP} \\
&\cite{helbing1995social}& \cite{van2011reciprocal}& \cite{han2025neupan} & \cite{cheng2024difftune} &\\
\midrule
-&\multicolumn{5}{c}{30 obstacles}\\
\addlinespace[1pt]
AvgSpeed\,$\uparrow$ &2.299 &1.938 &3.018 &\cellcolor{purple2}\textbf{3.023} &2.870\\
AvgAcc\,$\downarrow$ &1.015 &0.828 &\cellcolor{purple2}\textbf{0.410} &0.479 &0.774\\
AvgEpoch\,$\downarrow$ &-- &-- &-- &3.610 &\cellcolor{purple2}\textbf{2.750}\\
TotTime\,$\downarrow$ &35.538 &35.556 &\cellcolor{purple2}\textbf{19.108} &35.079 &38.453\\
ColRate\,$\downarrow$ &3.0\% &\cellcolor{purple2}\textbf{0.0\%} &22.0\% &17.0\% &1.0\%\\
SucRate\,$\uparrow$ &35.0\% &46.0\% &60.0\% &79.0\% &\cellcolor{purple2}\textbf{95.0\%}\\
\midrule
-&\multicolumn{5}{c}{40 obstacles}\\
\addlinespace[1pt]
AvgSpeed\,$\uparrow$ &2.251 &1.844 &\cellcolor{purple2}\textbf{3.065} &3.030 &2.944\\
AvgAcc\,$\downarrow$ &1.151 &0.840 &\cellcolor{purple2}\textbf{0.467} &0.568 &0.765\\
AvgEpoch\,$\downarrow$ &-- &-- &-- &3.660 &\cellcolor{purple2}\textbf{2.950}\\
TotTime\,$\downarrow$ &36.109 &37.620 &\cellcolor{purple2}\textbf{18.076} &34.359 &40.041\\
ColRate\,$\downarrow$ &6.0\% &\cellcolor{purple2}\textbf{0.0\%} &24.0\% &21.0\% &4.0\%\\
SucRate\,$\uparrow$ &21.0\% &29.0\% &56.0\% &76.0\% &\cellcolor{purple2}\textbf{91.0\%}\\
\midrule
-&\multicolumn{5}{c}{50 obstacles}\\
\addlinespace[1pt]
AvgSpeed\,$\uparrow$ &2.274 &1.942 &\cellcolor{purple2}\textbf{3.036} &2.988 &2.808\\
AvgAcc\,$\downarrow$ &1.222 &0.903 &\cellcolor{purple2}\textbf{0.529} &0.568 &0.959\\
AvgEpoch\,$\downarrow$ &-- &-- &-- &5.420 &\cellcolor{purple2}\textbf{4.080}\\
TotTime\,$\downarrow$ &37.988 &38.989 &\cellcolor{purple2}\textbf{16.437} &40.178 &49.440\\
ColRate\,$\downarrow$ &8.0\% &\cellcolor{purple2}\textbf{0.0\%} &29.0\% &35.0\% &5.0\%\\
SucRate\,$\uparrow$ &6.0\% &10.0\% &43.0\% &60.0\% &\cellcolor{purple2}\textbf{87.0\%}\\
\bottomrule
\end{tabular}
\label{Tab1}
\vspace{-0.1in}
\end{table}

We construct simulation scenarios in IR-SIM using a
$65\,\mathrm{m}\times65\,\mathrm{m}$ workspace and a differential-drive robot of size
$2.0\,\mathrm{m}\times1.6\,\mathrm{m}$.
We consider three obstacle densities containing 30, 40, and 50 obstacles, respectively.
For each density, 100 evaluation scenes are generated and shared, resulting in 300 evaluation cases per method.
As illustrated in Fig.~\ref{fig3}, the benchmark contains four scenario profiles that induce different planner-adaptation challenges.
The \emph{reference-conflict} profile introduces locally unreliable segments along an otherwise prescribed reference path.
The \emph{high-speed slalom} profile combines abrupt path curvature with high reference speeds.
The \emph{compound} profile combines these two sources of difficulty within the same navigation task.
Finally, the \emph{uniform-random} profile distributes irregular polygonal obstacles throughout the workspace and evaluates adaptation in unstructured clutter.
Together, these profiles cover both specifically designed failure patterns and randomized environments for evaluations. The maximum replanning epoch threshold $\mathcal{Y}$ is set to 10.

Fig.~\ref{fig4} presents a 50-obstacle uniform-random case, where the robot observes only a local portion of the environment during execution.
SFM and ORCA fail to make sufficient progress and eventually time out, while NeuPAN becomes stuck in the cluttered region.
DiffTune performs several gradient-based updates but ultimately collides with an obstacle. In contrast,  after two consecutive ineffective AD updates, the trigger in Eq.~\eqref{trigger} activates IL at epoch~3 instead of performing another local AD update, where \texttt{Qwen3-VL-Plus} interprets the recent adaptation history and identifies overly strong path and speed tracking as a likely cause of ineffective recovery. Thus, the planner is revised through reducing $q_s$, $p_u$, and $\eta$, increasing the obstacle-related loss weight $\gamma$, and adjusting $(d_{\min},d_{\max})$.
Moreover, the loss weights change from
$(0.25,0.25,0.25,0.25)$
to $(0.171,0.220,0.366,0.244)$. After that, SERP successfully traverses the cluttered region and reaches the goal. From this experiment, the following insight is obtained: \emph{Unlike AD, IL
enables rethinking from a whole new replanning direction, highlighting the superiority of reasoning-based tuning over
gradient-only tuning.}

\begin{figure*}[t!]
\centering
\includegraphics[width=0.95\textwidth]{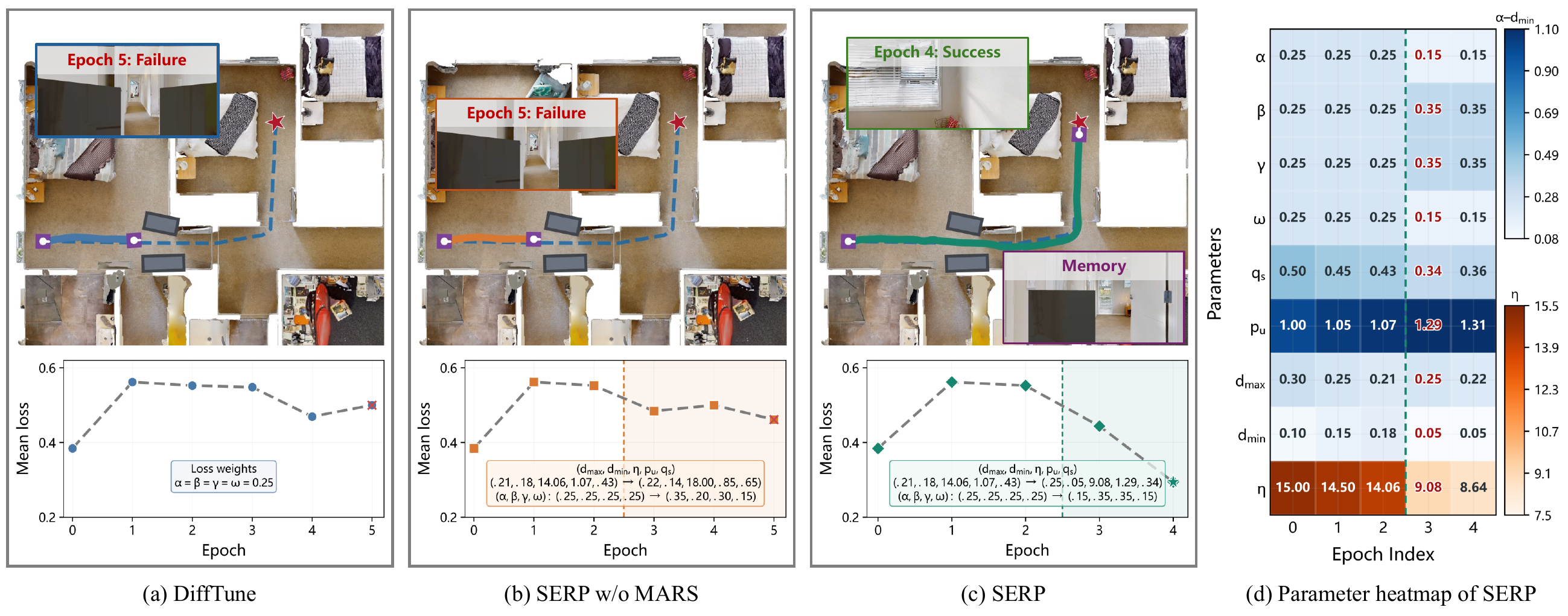} 
\vspace{-0.03in}
\caption{Qualitative comparison in Habitat simulator.}
\vspace{-0.12in}
\label{fig5}
\end{figure*}

Moreover, as shown in Table~\ref{Tab1}, the quantitative comparison is conducted over 300 evaluation cases, where SERP consistently achieves the highest success rate across all obstacle densities, reaching 95.0\%, 91.0\%, and 87.0\% with 30, 40, and 50 obstacles, respectively. The advantage becomes particularly evident in the densest setting, where SERP improves the success rate from 60.0\% for DiffTune to 87.0\% while reducing the collision rate from 35.0\% to 5.0\%.
SERP also requires fewer adaptation epochs than DiffTune at all three obstacle densities, indicating that ILAD can avoid repeatedly following ineffective local gradient directions.

\begin{table}[t!] \small
\setlength{\tabcolsep}{0.5mm}
\renewcommand{\arraystretch}{1.32}
\centering
\caption{Ablation study of SERP. \colorbox{purple2}{Purple} denotes the \textbf{best} performance.}
\begin{tabular}{cccccc}
\toprule
\multicolumn{2}{c}{\textbf{Method}} 
& AvgEpoch\,$\downarrow$ & TotTime\,$\downarrow$ & ColRate\,$\downarrow$ & SucRate\,$\uparrow$ \\
\midrule
\multicolumn{2}{c}{SERP} & \cellcolor{purple2}\textbf{2.66} & 43.75 & \cellcolor{purple2}\textbf{2.0\%} & \cellcolor{purple2}\textbf{94.0\%} \\
\midrule
\multicolumn{2}{c}{SERP w/o IL} & 3.82 & \cellcolor{purple2}\textbf{43.66} & 20.0\% & 74.0\% \\
\midrule
\multicolumn{2}{c}{SERP w/o AD} & 3.22 & 48.47 & 4.0\% & 84.0\% \\
\bottomrule
\end{tabular}
\label{Tab2}
\end{table}

Table~\ref{Tab2} further evaluates the complementary contributions of IL and AD on 100 scenes with 50 obstacles, using a separate set of random seeds from those employed in Table~\ref{Tab1}. Removing IL decreases the success rate from 94.0\% to 74.0\% and increases the collision rate from 2.0\% to 20.0\%.
This result shows that local differentiable adaptation alone can remain constrained to an ineffective parameter--loss region. Moreover, removing AD reduces the success rate to 84.0\% and increases AvgEpoch from 2.66 to 3.22, indicating that semantic intervention alone lacks the efficient local refinement provided by gradient-based optimization. These results support the complementary roles of IL and AD: \emph{IL helps AD to escape ineffective configuration region, whereas AD reshapes the coarse understanding of IL into detailed fine-grained parameter differentiation.}

\subsection{Habitat Simulation}

We further evaluate SERP in six photorealistic indoor scenes from the
HM3D dataset~\cite{ramakrishnan2021hm3d} including \texttt{00804}, \texttt{00805},
\texttt{00806}, \texttt{00807},
\texttt{00812}, and \texttt{00813}, each with 50 generated scenarios. Specifically, start and goal are
sampled from navigable locations on the same floor, and the shortest path in the original scene is used as the
reference path. To create local-planning challenges, a geometry-constrained generator places near-path obstacles, narrow passages, corner obstacles, or partial blockages at sampled locations along the path guided by VLM. After obstacle
insertion, we retain only scenarios in which a feasible route remains close to the original reference path, thereby excluding cases that require global rerouting. For all Habitat evaluations, the recovery memory is initialized as empty, and the evaluation scenarios are randomly shuffled. After each episode, only the recovery trace obtained from the completed real execution is consolidated into memory and becomes available to subsequent episodes. Therefore, the memory is accumulated strictly online, without access to any future evaluation scenarios.

\begin{table}[t!] \small
\setlength{\tabcolsep}{0.5mm}
\renewcommand{\arraystretch}{1.32}
\centering
\caption{Quantitative comparison in the Habitat simulator. \colorbox{purple2}{Purple} denotes the \textbf{best} performance.}
\begin{tabular}{cccccc}
\toprule
\multicolumn{2}{c}{\textbf{Method}} & AvgEpoch$\downarrow$ & TotTime$\downarrow$ & ColRate$\downarrow$ & SucRate$\uparrow$ \\
\midrule
\multicolumn{2}{c}{SERP} & \cellcolor{purple2}\textbf{4.06} & 74.72 & \cellcolor{purple2}\textbf{1.7\%} & \cellcolor{purple2}\textbf{87.7\%} \\
\midrule
\multicolumn{2}{c}{DiffTune \cite{cheng2024difftune}} & 4.48 & \cellcolor{purple2}\textbf{49.46} & 4.7\% & 77.0\% \\
\midrule
\multicolumn{2}{c}{SERP w/o MARS} & 4.35 & 55.90 & 2.7\% & 81.3\% \\
\bottomrule
\end{tabular}
\label{Tab3}
\end{table}

\begin{figure}[t]
  \centering
  \begin{subfigure}[t]{0.45\textwidth}
    \centering
    \includegraphics[width=0.92\textwidth]{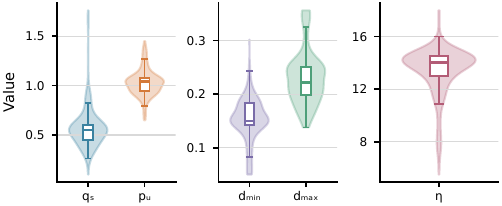}
    \caption{Parameter distributions of successful recoveries.}
  \end{subfigure}
  \begin{subfigure}[t]{0.45\textwidth}
    \centering
    \includegraphics[width=0.92\textwidth]{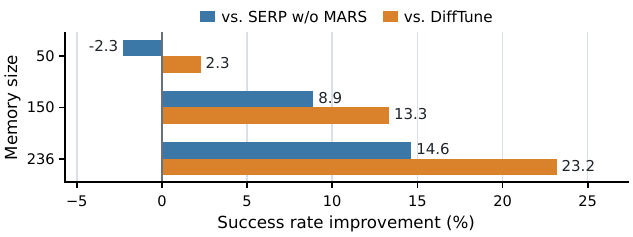}
    \caption{Advantage of SERP over baseline schemes.}
  \end{subfigure}
  \vspace{-0.02in}
  \caption{Analysis of SERP in the Habitat simulator.}
  \label{fig6}
   \vspace{-0.05in}
\end{figure}

Fig.~\ref{fig5} compares different
adaptation strategies in a narrow-passage case in scenario \texttt{00804}. As shown in Fig.~\ref{fig5}(a)--(b), neither DiffTune
nor SERP without MARS reaches the goal. In contrast, SERP invokes MARS at epoch~3 to retrieve relevant memory entries as shown in Fig.~\ref{fig5}(c). The retrieved memory depicts failures with similar narrow-passage structure, where the robot becomes stuck around the bottleneck. Then \texttt{Qwen3.7-Plus} generate four alternative recovery
configurations. Each candidate is
evaluated through a low-fidelity rollout in IR-SIM comprising at
most $N=3$ rollouts. As shown in
Fig.~\ref{fig5}(d), MARS selects
\texttt{cand\_relax\_tracking}, which decreases $q_s$ and $\eta$,
increases $p_u$, changes the loss weights from
$(0.25,0.25,0.25,0.25)$ to $(0.15,0.35,0.35,0.15)$, and widens
$(d_{\min},d_{\max})$ from $[0.18,0.21]\,\mathrm{m}$ to
$[0.05,0.253]\,\mathrm{m}$. Due to the fidelity gap between IR-SIM and Habitat, the selected configuration does not immediately complete the recovery after transfer. Nevertheless, it provides a promising configuration initialization, from which one additional adaptation update enables the robot to pass the narrow passage and reach the goal, as shown in
Fig.~\ref{fig5}(c). This case provides the following insight: \emph{MARS enhances failure recovery by leveraging memory to generate high quality candidate configurations and rollout-based evaluation to identify promising adaptation directions.}

Table~\ref{Tab3} reports the quantitative results in Habitat.
SERP achieves the highest success rate of 87.7\%, compared with
77.0\% for DiffTune and 81.3\% for SERP without MARS, while also
requiring fewer adaptation epochs. Fig.~\ref{fig6}
provides further insight into this improvement. Successful recoveries
span diverse planner parameters as shown in Fig.~\ref{fig6}(a), indicating that complex indoor
environments may admit multiple effective recovery strategies.
Moreover, Fig.~\ref{fig6}(b) examines how recovery performance evolves as memory is accumulated online during evaluation. As the memory size grows from 50 to 236, the success rate advantage of SERP over SERP without MARS and DiffTune increases from -2.3\%/2.3\% to 14.6\%/23.2\%, respectively. This trend suggests that accumulated recovery memory provides increasingly informative guidance for generating and selecting effective configurations.

\begin{figure}[!t]
    \centering
    \begin{subfigure}{0.45\linewidth}
        \centering
        \includegraphics[width=\linewidth]{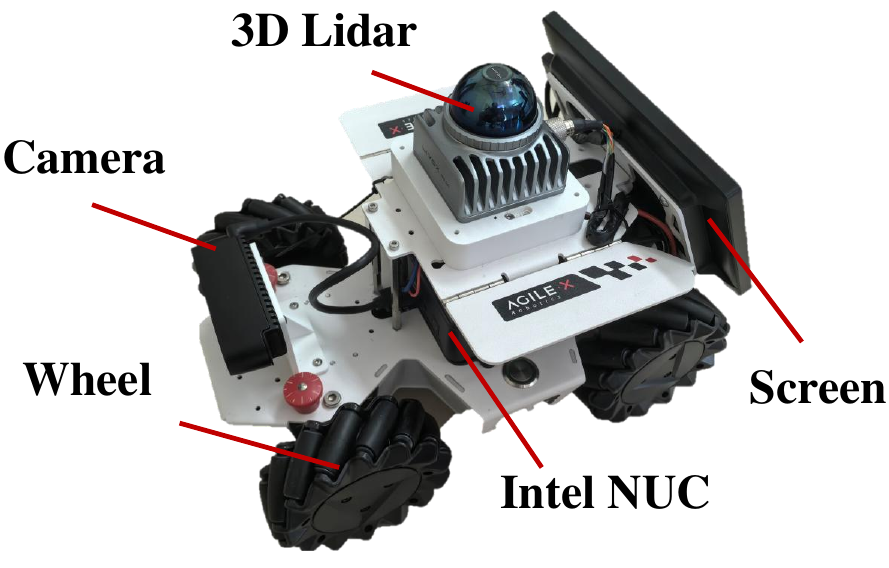}
        \caption{Robot.}
    \end{subfigure}
    \hfill
        \begin{subfigure}{0.261\linewidth}
        \centering
        \includegraphics[width=\linewidth]{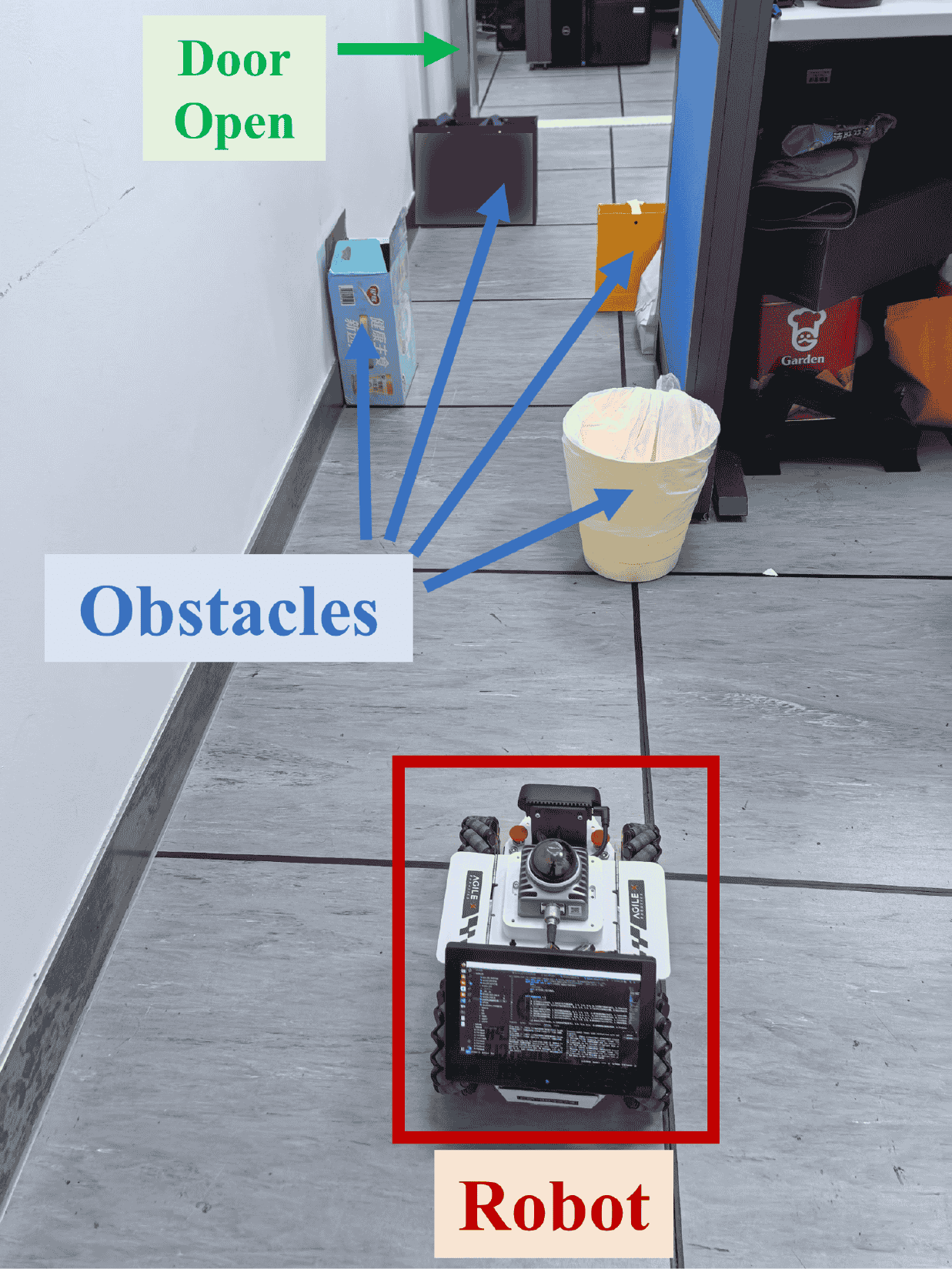}
        \caption{Case 1.}
    \end{subfigure} 
            \hfill
    \begin{subfigure}{0.261\linewidth}
        \centering
        \includegraphics[width=\linewidth]{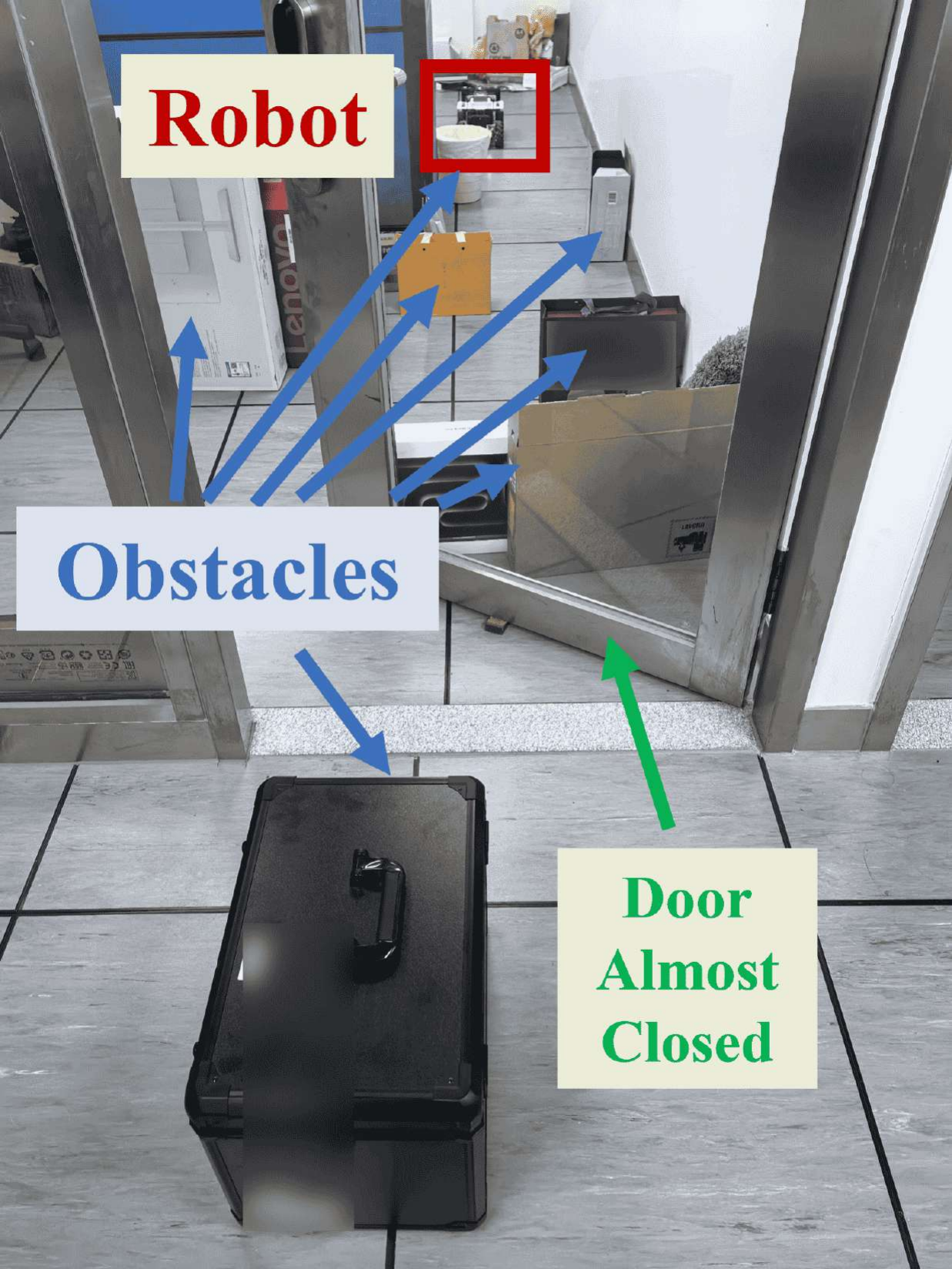}
        \caption{Case 2.}
    \end{subfigure} 
    \vspace{-0.2in}
    \caption{Experimental setup.}
    \vspace{-0.05in}
    \label{fig7}
\end{figure}

\subsection{Real-World Experiments}

We further evaluate SERP on a physical car-like robot to examine the feasibility of online planner adaptation under real execution conditions.
The robot is equipped with a 3D lidar, an RGB-D camera, and an Intel NUC running the SLAM and planning modules, as illustrated in Fig.~\ref{fig7}a.
The robot starts inside an office room, as shown in Fig.~\ref{fig7}b--c, and navigates through a constrained doorway with additional obstacles placed around the passage.
Two door configurations are considered: \emph{open} and \emph{almost closed}.

\begin{figure}[t]
  \centering
  \begin{subfigure}[t]{0.45\textwidth}
    \centering
    \includegraphics[width=0.92\textwidth]{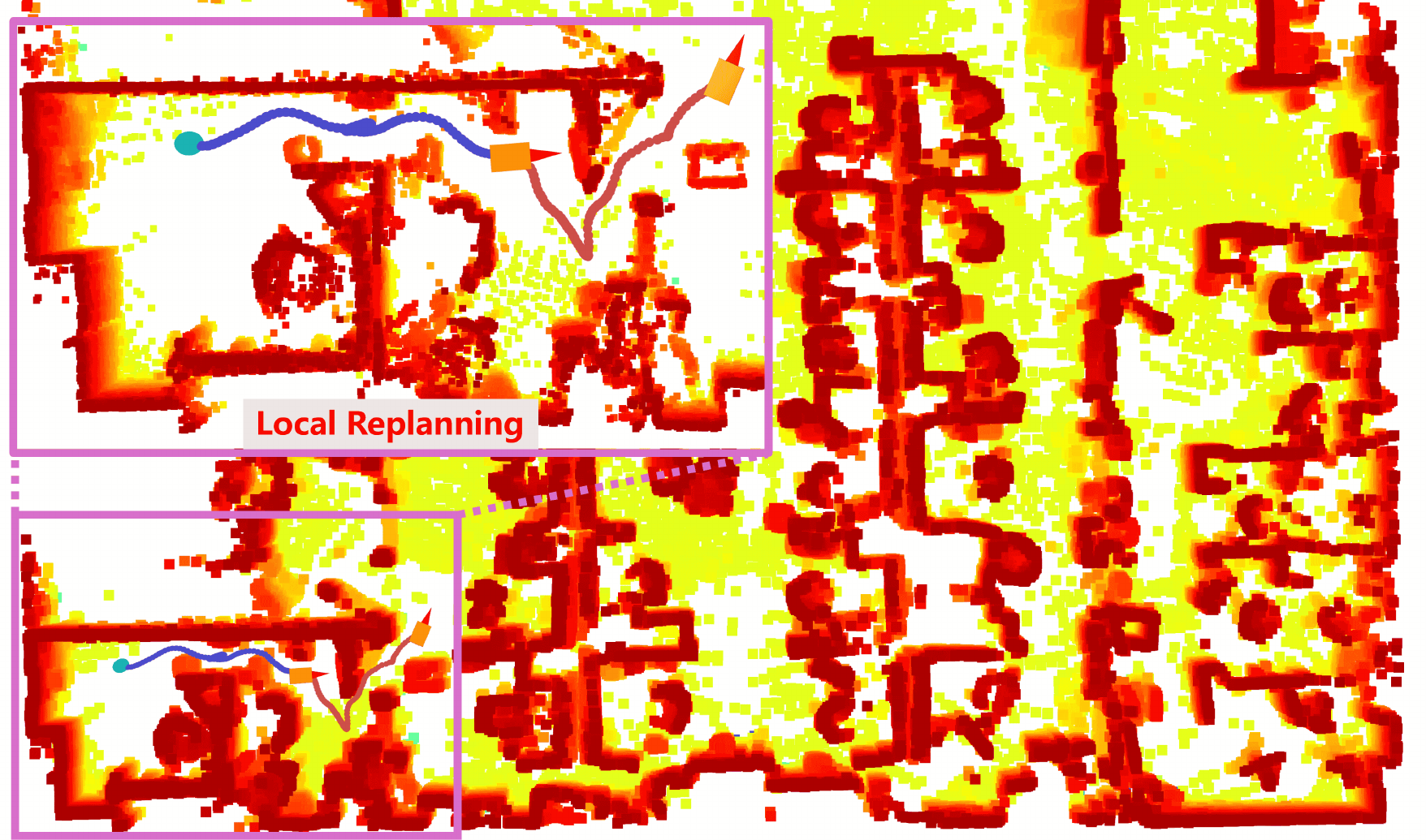}
    \caption{ Trajectories in the point cloud. }
    \label{limo_carto}
  \end{subfigure}
  \begin{subfigure}[t]{0.45\textwidth}
    \centering
    \includegraphics[width=0.92\textwidth]{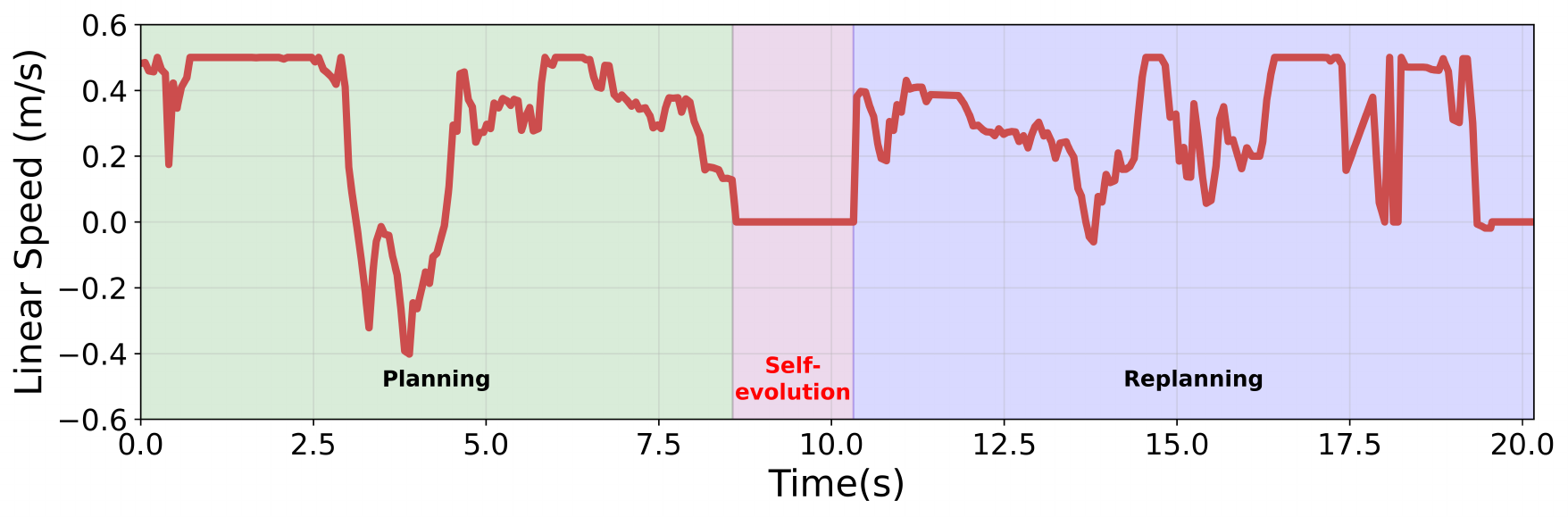}
    \caption{ Linear speed }
    \label{limo_linear}
  \end{subfigure}
  \begin{subfigure}[t]{0.45\textwidth}
    \centering
    \includegraphics[width=0.92\textwidth]{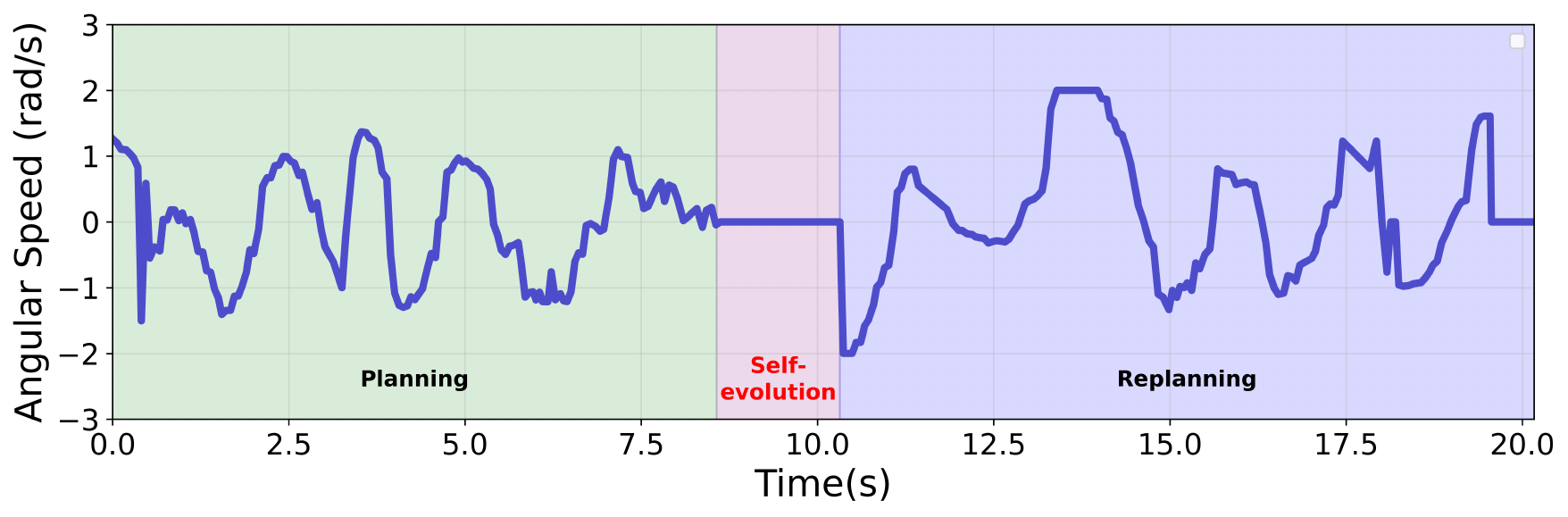}
    \caption{ Angular speed }
    \label{limo_angular}
  \end{subfigure}
  \vspace{-0.08in}
  \caption{Trajectory and control profiles of SERP. }
  \vspace{-0.05in}
  \label{fig8}
\end{figure}

\begin{figure*}[t!]
\centering
\includegraphics[width=0.95\textwidth]{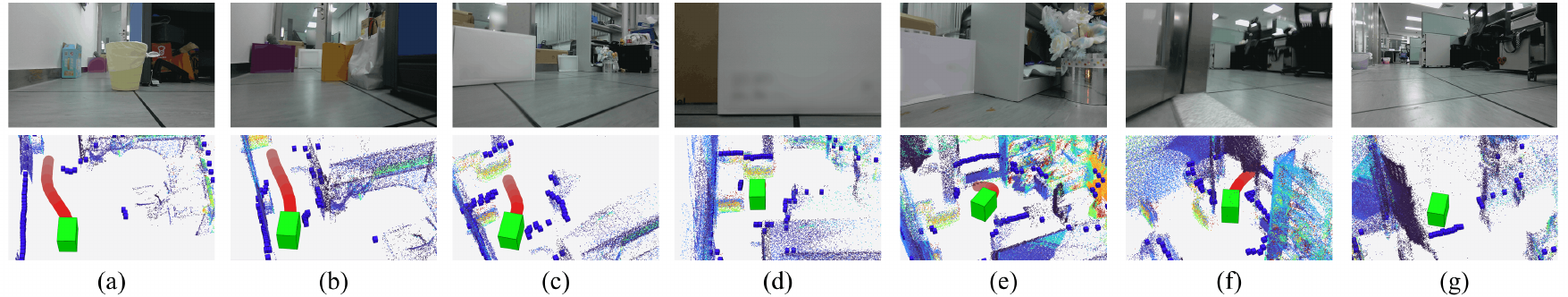} 
\caption{Replanning in the cluttered laboratory.}
\vspace{-0.1in}
\label{fig9}
\end{figure*}

SERP is evaluated in Case~2 in Fig.~\ref{fig7}c, where the planner parameters
$\mathcal{\theta}=\{q_s,p_u,\eta\}=\{0.30,0.80,29.00\}$
is used to initialize navigation in the test environment.
The resulting trajectory and control profiles are shown in Fig.~\ref{fig8}, while the detailed navigation process is illustrated in Fig.~\ref{fig9}.
The robot becomes stuck in front of the narrow doorway at
$t=8.57\,\mathrm{s}$.
After detecting the failure, SERP performs ILAD and updates the configuration to
$\mathcal{\theta}=\{0.12,0.62,29.18\}$
within one adaptation epoch and less than $3\,\mathrm{s}$ of computation.
The revised configuration enables the robot to traverse the doorway at
$t=10.32\,\mathrm{s}$ and continue navigation.

\begin{table*}[t]
\centering
\caption{Ablation study of SERP in real experiments.}
\resizebox{\textwidth}{!}{%
\begin{tabular}{ccccccc@{\hspace{1em}}cc@{\hspace{1em}}cc}
\toprule
\multirow{2}{*}{Door}
& \multirow{2}{*}{\# Obstacles}
& \multicolumn{2}{c}{Success}
& \multicolumn{2}{c}{Navigation Time (s)}
& \multicolumn{2}{c}{Speed ~Avg./Std (m/s)}
& \multicolumn{2}{c}{Acc.~Avg./Std (m/s\textsuperscript{2})} \\
\cmidrule(lr){3-4}
\cmidrule(lr){5-6}
\cmidrule(lr){7-8}
\cmidrule(lr){9-10}
& & w/ SERP & w/o SERP
& w/ SERP & w/o SERP
& w/ SERP & w/o SERP
& w/ SERP & w/o SERP \\
\midrule
Open
& 2
& \CheckmarkBold & \CheckmarkBold
& \textbf{10.78} & 11.13
& \textbf{0.415 / 0.166} & 0.396 / 0.179
& \textbf{0.487 / 1.587} & 0.636 / 1.875 \\

Open
& 4
& \CheckmarkBold & \CheckmarkBold
& \textbf{10.97} & 11.46
& \textbf{0.419 / 0.151} & 0.392 / 0.172
& 0.484 / 1.218 & \textbf{0.418 / 0.899} \\

Open
& 6
& \CheckmarkBold & \CheckmarkBold
& \textbf{11.42} & 11.70
& \textbf{0.416 / 0.131} & 0.405 / 0.180
& 0.537 / \textbf{0.988} & \textbf{0.485} / 1.625 \\

Almost Closed
& 2
& \CheckmarkBold & \XSolidBrush
& \textbf{19.55} & --
& \textbf{0.274 / 0.212} & --
& \textbf{0.405 / 1.281} & -- \\

Almost Closed
& 4
& \CheckmarkBold & \XSolidBrush
& \textbf{14.46} & --
& \textbf{0.378 / 0.151} & --
& \textbf{0.651 / 1.754} & -- \\

Almost Closed
& 6
& \CheckmarkBold & \XSolidBrush
& \textbf{21.76} & --
& \textbf{0.263 / 0.203} & --
& \textbf{0.787 / 1.897} & -- \\
\bottomrule
\end{tabular}%
}
\label{Tab4}
\vspace{-0.06in}
\end{table*}

Table~\ref{Tab4} further evaluates the effect of SERP with ILAD enabled under different doorway constraints.
When the door is open, both variants successfully traverse the passage for all three obstacle configurations, with similar average navigation times of $11.06\,\mathrm{s}$ with SERP and $11.43\,\mathrm{s}$ without SERP. With SERP enabled, the robot successfully traverses the doorway in all three tested obstacle configurations, with an average navigation time of $18.59\,\mathrm{s}$.
Without SERP, the robot fails to pass the constrained doorway in all three cases.
These results show that SERP becomes beneficial when the encountered geometry requires planner parameters different from the nominal one.

\vspace{-0.05in}
\section{Conclusion}\label{section7}

This paper introduced SERP, a self-evolutionary replanning framework for failure-aware motion planning.
ILAD combines history-informed semantic intervention with auto-differentiation to reshape planner adaptation when local numerical adaptation becomes ineffective.
MARS further extends this process through memory-guided candidate generation and parallel low-fidelity rollout evaluation, while consolidating executed recovery traces into memory to guide future adaptation.
Together, these mechanisms enable SERP to recover from diverse planner failures through semantic intervention, numerical optimization, and accumulated memory.
Extensive experiments in IR-SIM, Habitat, and real-world environments demonstrate improved failure-recovery performance over various baselines, while ablation studies verify the complementary roles of IL and AD and the benefit of MARS.

\vspace{-0.05in}
\bibliographystyle{IEEEtran}
\bibliography{reference/Thesis}

\end{document}